%% file: acl_latex.tex
\pdfoutput=1

\documentclass[11pt]{article}

\usepackage[final]{acl}

\usepackage{times}
\usepackage{latexsym}
\usepackage{lipsum}

\usepackage[T1]{fontenc}

\usepackage[utf8]{inputenc}

\usepackage{microtype}

\usepackage{tabularx}
\usepackage{textcomp}
\usepackage{listings}
\usepackage{inconsolata}

\usepackage{graphicx}

\usepackage{booktabs}
\usepackage{hyperref}
\usepackage{tablefootnote}
\usepackage{tcolorbox}

\definecolor{models_color}{HTML}{4285f4} 
\definecolor{tasks_color}{HTML}{fbbc05} 
\definecolor{format_color}{HTML}{34a853} 
\definecolor{metrics_color}{HTML}{673ab7} 
\definecolor{results_color}{HTML}{ea4335} 

\newcommand{\coloredsquare}[1]{\textcolor{#1}{$\blacksquare$}}

\usepackage{amssymb}
\usepackage{arydshln}
\makeatletter
\def\adl@drawiv#1#2#3{%
        \hskip.5\tabcolsep
        \xleaders#3{#2.5\@tempdimb #1{1}#2.5\@tempdimb}%
                #2\z@ plus1fil minus1fil\relax
        \hskip.5\tabcolsep}
\newcommand{\cdashlinelr}[1]{%
  \noalign{\vskip 1.3pt
           \global\let\@dashdrawstore\adl@draw
           \global\let\adl@draw\adl@drawiv}
  \cdashline{#1}[.4pt/2pt]
  \noalign{\global\let\adl@draw\@dashdrawstore
           \vskip 3pt}}
\makeatother

\title{\textsc{MT-Lens}: An all-in-one Toolkit for Better Machine Translation Evaluation }

\author{Javier García Gilabert, Carlos Escolano, Audrey Mash, Xixian Liao, Maite Melero \\
Barcelona Super Computing Center (BSC) \\
{\small \tt \{javier.garcia1,carlos.escolano,} \\
{\small \tt audrey.mash,xixian.liao,maite.melero\}}{\small \tt @bsc.es}
}

\newcommand{\mtlens}{\textsc{MT-Lens}}

\newcommand{\bleu}{\small \textsc{BLEU}}
\newcommand{\ter}{\small \textsc{TER}}
\newcommand{\chrf}{\small \textsc{ChrF}}

\newcommand{\comet}{\small \textsc{COMET}}
\newcommand{\bleurt}{\small \textsc{BLEURT}}
\newcommand{\metricx}{\small \textsc{MetricX}}
\newcommand{\xcomet}{\small \textsc{Xcomet}}
\newcommand{\xcometqe}{\small \textsc{Xcomet-QE}}
\newcommand{\metricxqe}{\small \textsc{MetricX-QE}}
\newcommand{\cometkiwi}{\small \textsc{Comet-Kiwi}}

\newcommand{\etox}{\small \textsc{ETOX}}
\newcommand{\mutox}{\small \textsc{MuTox}}
\newcommand{\detoxify}{\small \textsc{Detoxify}}

\newcommand{\holisticbias}{\textsc{HolisticBias\;}}
\newcommand{\floresdataset}{\textsc{FLORES-200\;}}

\definecolor{codegreen}{rgb}{0,0.6,0}
\definecolor{codegray}{rgb}{0.5,0.5,0.5}
\definecolor{codepurple}{rgb}{0.58,0,0.82}
\definecolor{backcolour}{rgb}{0.95,0.95,0.92}

\lstdefinestyle{mystyle}{
    backgroundcolor=\color{backcolour},   
    commentstyle=\color{codegreen},
    keywordstyle=\color{magenta},
    numberstyle=\tiny\color{codegray},
    stringstyle=\color{codepurple},
    basicstyle=\ttfamily\footnotesize,
    breakatwhitespace=false,         
    breaklines=true,                 
    captionpos=b,                    
    keepspaces=true,                 
    numbers=left,                    
    numbersep=5pt,                  
    showspaces=false,                
    showstringspaces=false,
    showtabs=false,                  
    tabsize=2
}

\begin{document}
\maketitle
\begin{abstract}

We introduce {\mtlens}\footnote{We release our code at \url{https://github.com/langtech-bsc/mt-evaluation}. Demo is available at this \href{https://ezjg83tbznvq9ue6zxiaxj.streamlit.app/}{link} while the demo video is available at this \href{https://www.youtube.com/watch?v=6D6CUQyLGpY}{link}. }, a framework designed to evaluate Machine Translation (MT) systems across a variety of tasks, including translation quality, gender bias detection, added toxicity, and robustness to misspellings. While several toolkits have become very popular for benchmarking the capabilities of Large Language Models (LLMs), existing evaluation tools often lack the ability to thoroughly assess the diverse aspects of MT performance. {\mtlens} addresses these limitations by extending the capabilities of LM-eval-harness for MT, supporting state-of-the-art datasets and a wide range of evaluation metrics. It also offers a user-friendly platform to compare systems and analyze translations with interactive visualizations. {\mtlens} aims to broaden access to evaluation strategies that go beyond traditional translation quality evaluation, enabling researchers and engineers to better understand the performance of a NMT model and also easily measure system's biases.

\end{abstract}

\section{Introduction}

Neural Machine Translation (NMT) models are typically evaluated using automated metrics that compare the model's translated outputs to one or more reference translations. Recent neural-based evaluation metrics have demonstrated high correlations with human judgments \cite{rei-etal-2022-comet, sellam-etal-2020-bleurt, juraska-etal-2023-metricx} replacing  canonical overlap-based metrics \cite{popovic-2015-chrf, papineni-etal-2002-bleu}. While automated metrics provide an essential means for assessing quality improvements and have become the way to go in evaluating state-of-the-art NMT systems, they only offer a general intuition of the model’s overall performance and often lack interpretability \cite{perrella-etal-2024-guardians}. As such, the reliance on these metrics raises concerns about their specific error types that might affect the end-user experience, such as reinforcing gender bias  \cite{zaranis2024watching}, translationese or language mismatch \cite{zouhar2024pitfalls}. 

To address these problems, recent works have focused on developing inherently interpretable metrics for MT evaluation \cite{guerreiro2023xcomet} via token-level annotations that offer a more granular-insight at the segment level. However, such limitations necessitate more granular evaluation tools that can dissect translation outputs to identify and analyze specific error types and their impact on overall quality, thereby enabling MT engineers to make more informed decisions when evaluating translation systems.

Moreover, existing evaluation methodologies in state-of-the-art NMT models primarily focus on evaluating translation quality, frequently overlooking other equally critical evaluations like gender bias, added toxicity or robustness to misspellings. These biases and harmful outputs can have significant consequences for users \cite{savoldi2024harm}, highlighting a need for evaluation strategies that go beyond quality to encompass a broader range of tasks.

\mtlens\; seeks to address these critical gaps by providing a unified framework to test generative language models on a number of different machine translation evaluation tasks and providing a user-friendly interface to analyze the results. \mtlens\; is based on LM-eval-harness \cite{eval-harness} which has been widely used for evaluating LLMs in several Natural Language Understanding (NLU) tasks. Building upon this comprehensive framework, \mtlens\; extends the evaluation capabilities of LM-eval-harness for NMT. Contributions of this framework are listed as follows:


\input{latex/tables/datasets}

\begin{itemize}
    \item We support novel evaluation tasks for detecting gender bias, added toxicity, and robustness to character noise in MT.

    \item We support a variety of state-of-the-art benchmark datasets and evaluation metrics for assessing translation quality.

    \item We provide a user-friendly interface with interactive visualizations at both segment and system levels, enabling thorough error analysis and performance assessment of evaluations performed using \mtlens.

    \item We provide bootstrapping significance tests for comparing MT systems on both neural-based and overlap-based machine translation metrics.
    
\end{itemize}

{\mtlens} is designed to broaden access to novel evaluation tasks that go beyond traditional translation quality in MT, while also supporting general NLU tasks. The framework is maintained by the Language Technologies Unit at the Barcelona Supercomputing Center, ensuring ongoing updates with the latest features of LM-eval-harness and support for new MT datasets developed by the research community.

\section{Related work}

Typically, automatic metrics like \textsc{Bleu} \citep{papineni-etal-2002-bleu} are simply applied and reported at the corpus level. In recent years, the evolution of MT evaluation has seen the development of tools that offer more granular insights.

\textsc{MT-CompareEval} \citep{klejch2015mt} provides comparative analysis of segment-level errors, focusing on identifying differences in n-grams between two MT outputs.
Similarly, \textsc{Compare-MT} \citep{neubig-etal-2019-compare} offers a holistic analysis for pairs of MT systems, examining performance metrics such as n-gram frequency and part-of-speech accuracy. \textsc{MATEO} \cite{vanroy-etal-2023-mateo} offers a friendly web-based interface for evaluating MT outputs on several evaluation metrics. 
\textsc{MT-Telescope}, a newer platform developed by \citet{rei-etal-2021-mt}, not only supports segment-level analysis but also offers a web-based interface for better visualization of comparative performance between two MT systems. In addition, it enables focused analysis on phenomena like named entity translation and terminology handling. The last three mentioned tools include statistical significance testing through bootstrapped t-tests to ensure the reliability and statistical validity of their evaluations. Importantly, these tools require users to upload the source sentences and system outputs to perform evaluations.

More recently, \textsc{TowerEval} \citep{tower_llm_2024} has been developed specifically for LLMs. This evaluation framework allows users to benchmark their models against a comprehensive suite of datasets for evaluating translation quality.
It supports both generation and evaluation processes, allowing users to run inference and compute a variety of metrics such as \textsc{Bleu}, \textsc{Comet} \citep{rei-etal-2022-comet}, \textsc{Comet-Kiwi} \citep{rei-etal-2022-cometkiwi}, \textsc{ChrF} \citep{popovic-2015-chrf} and \textsc{TER} \citep{snover-etal-2006-study}. 
It also allows for the creation of custom test suites and instructions. 

Although the aforementioned works aimed to become the standard tools for evaluating NMT systems, they have not achieved widespread adoption within the MT research community. On the contrary, the LM-eval-harness library has gained significant popularity in the NLP community due to its extensibility, modularity, and ease of use. By building on this widely adopted framework, {\mtlens} bridges the gap between the versatility of LM-eval-harness and the specific needs of MT evaluation, thereby addressing the limitations of prior works.

{\mtlens} is largely inspired by previous frameworks but differentiates itself by being a unified framework where the user can easily run evaluations on the desired model and visualize the results in a user-friendly interface. It also supports a broader spectrum of MT tasks, including bias detection, toxicity evaluation, and robustness to character noise, in addition to traditional translation quality evaluation.



\input{latex/tables/metrics}

\section{Building blocks}


{\mtlens} follows a similar evaluation methodology as LM-eval-harness for MT tasks. When evaluating a system, it follows a predefined sequence of steps which can be divided into five main blocks: \coloredsquare{models_color} Models, \coloredsquare{tasks_color} Tasks, \coloredsquare{format_color} Format, \coloredsquare{metrics_color} Metrics and \coloredsquare{results_color} Results. Once the evaluation is completed the user interface will show a fine-grained analysis of the system with interactive visualizations.\\

\noindent
\coloredsquare{models_color} \textbf{Models} {\mtlens} supports different inference frameworks for running MT tasks: \texttt{fairseq} \cite{ott2019fairseq}, \texttt{CTranslate2}\footnote{\href{https://github.com/OpenNMT/CTranslate2}{https://github.com/OpenNMT/CTranslate2}}, \texttt{transformers} \cite{wolf-etal-2020-transformers}, \texttt{vllm} \cite{kwon2023efficient}, and others. If a model is not directly supported, users can utilize the \texttt{simplegenerator} wrapper, which accepts pre-generated translations from a text file. \\


\noindent
\coloredsquare{tasks_color} \textbf{Tasks} A MT task is defined by the dataset to be used and the language pair involved (Table \ref{table:datasets}). Each MT task is uniquely named using the convention:

\medskip

\begin{center}
\small
\begin{verbatim}
  {source language}_{target language}_{dataset}
\end{verbatim}
\end{center}

\medskip

\noindent
\coloredsquare{format_color} \textbf{Format} When executing MT tasks we need to define how input prompts are formatted for the selected model. Different models may require the source sentence to be formatted in a specific style. Users can specify the desired template through a YAML file, and the model implementation will automatically format the source sentence accordingly.\\

\noindent
\coloredsquare{metrics_color} \textbf{Metrics} \mtlens\ includes an extensive number of evaluation metrics for MT tasks. These metrics cover both overlap-based and neural-based metrics which are listed in Table \ref{table:metrics}. Metrics are computed at the segment level and then aggregated at the system level. Each metric has some configurable hyper-parameters that users can adjust through a YAML file. \\

\noindent
\coloredsquare{results_color} \textbf{Results} Evaluation results are outputted in JSON format including source sentences, reference translations, aggregated metrics and segment level scores. Each JSON evaluation file is then used by the {\mtlens \; \textsc{UI} } to provide a more intuitive analysis for the user.

\subsection{Example usage}

Given an NMT model and a specified task, we can evaluate the model using {\mtlens} with the following command:

\small
\begin{lstlisting}
model='./models/madlad400/'
output_dir='results/results.json'

lm_eval --model hf \
  --model_args "pretrained=${model}" \
  --tasks en_ca_flores_devtest \
  --output_path $output_dir \
  --translation_kwargs "
                src_language=eng_Latn, 
                tgt_language=cat_Latn, 
                prompt_style=madlad400
                "
\end{lstlisting}
\normalsize

where the \texttt{hf} model argument indicates that the model is implemented using \texttt{transformers}. Then, we specify the model path, source and target languages, and prompt style. The task is set to \texttt{en\_ca\_flores\_devtest}, and the results will be saved to the specified output directory.

\subsection{MT Tasks}

In this section we outline the MT related tasks implemented in \mtlens.\\

\noindent
\textbf{General-MT} This task consists in evaluating the faithfulness and the quality of the translation using reference-based and quality estimation metrics. We show in Table \ref{table:datasets} the datasets that are natively supported in the \mtlens\; framework. \\ 

\noindent
\textbf{Added toxicity} This type of toxicity arises when a toxic element appears in the translated sentence without a corresponding toxic element in the source sentence, or when a toxic element in the translation results from a mistranslation of a non-toxic element in the source sentence. We use the \holisticbias
dataset \cite{smith-etal-2022-im} to evaluate NMT models on this task, which has previously been used for identifying added toxicity in NMT models \cite{garcia-gilabert-etal-2024-resetox, costa-jussa-etal-2024-added}. \holisticbias consists of approximately 472k sentences in English that are created using sentence templates across 13 demographic axes (gender, ability, religion, etc.). For measuring added toxicity we first filter the source sentences using {\mutox} \cite{costa-jussa-etal-2024-mutox} as done in \cite{tan2024massivemultilingualholisticbias} and measure the toxicity in the translations using {\etox} \cite{costa-jussa-etal-2023-toxicity}, {\mutox} \cite{costa-jussa-etal-2024-mutox} and {\detoxify} \cite{detoxify} toxicity classifiers. Since {\etox} supports 200 languages, it allows us to evaluate a wide range of languages for added toxicity using English as the source language. We then measure the translation faithfulness using {\cometkiwi} on the toxic sentences detected by each toxicity classifier as it has been proved useful to evaluate hallucinations when no reference is available \cite{guerreiro-etal-2023-optimal}. \\

\noindent
\textbf{Gender bias} This type of bias in translation occurs when the system's prediction is skewed toward a specific gender due to stereotypes or inequalities \cite{DBLP:conf/chi/FriedmanN95, DBLP:conf/eacl/SavoldiPFNB24, sant-etal-2024-power}. Additionally, as not all languages contain the same amount of gender information, when translating from a notionally gendered or ungendered language to a grammatically gendered language, decisions about gender assignation may need to be made from little or no context. This leads to gender bias in cases where gender is consistently assigned following stereotypical patterns, such as labelling all nurses as female and all doctors as male. 

We implement three tasks for the evaluation of gender bias when translating out of English, one using the \textsc{MuST-SHE} dataset \cite{bentivogli-etal-2020-gender, mash-etal-2024-unmasking} one using the Massive Multilingual Holistic Bias Dataset (\textsc{MMHB}) \cite{tan2024massivemultilingualholisticbias}, and finally, one using \textsc{MT-GenEval} \cite{currey-etal-2022-mtgeneval}.

\textsc{MuST-SHE} contains approximately 1000 English sentences. All English terms that must be assigned a gender in translation have been identified, and tuples containing both the correctly and incorrectly gendered translated forms are provided for evaluation. The use of transcripts of natural speech allows for the assessment of gender bias in more complex contextual clues and co-reference situations compared to template-based datasets. The accuracy of sex is measured using the revised script of \cite{mash-etal-2024-unmasking} and is reported at the sentence and data set level, allowing for a fine-grained analysis of model performance.

\textsc{MMHB} makes use of placeholder-based sentence generation to generate feminine, masculine and neutral variations of each sentence pattern  \cite{tan2024massivemultilingualholisticbias}. The dataset consists of 152,720 English sentences partitioned into train, dev and devtest splits. The placeholder-based system allows for the creation of sentences covering all variations of morphological agreement in the target languages. The created sentences are then organized into groupings and the {\chrf} scores are measured for the different subsets.

\textsc{MT-GenEval} consists of two distinct tasks, one operating at sentence level and the other looking at accuracy when extracting gender from the preceding context. In both cases, all data has been human-reviewed to exclude sentences with ambiguous gender references, and counterfactual data created. When dealing with contextual inputs, professions are grouped into stereotypically feminine, masculine and neutral following \citet{troles-schmid-2021-extending}. The results are gender-balanced datasets across 600 single sentences and 1100 contextual inputs, providing a measure of accuracy in translating gender. \\

\noindent
\textbf{Robustness to Character Noise}  
This task evaluates how introducing word-level synthetic errors into source sentences affects the translation quality of an NMT model. We utilize the \floresdataset devtest dataset \cite{nllbteam2022language}, which allows us to evaluate the model's robustness to character perturbations across a wide range of directions. We implement three types of synthetic noise that have been previously used to stress NMT systems \cite{belinkovsynthetic, peters2024did}:

\begin{itemize}
    \item \textbf{swap:} For a selected word, two adjacent characters are swapped.
    \item \textbf{chardupe:} A character in the selected word is duplicated. 
    \item \textbf{chardrop:} A character is deleted from the selected word. 
\end{itemize}

A noise level parameter $\lambda \in [0, 1]$ controls the proportion of words in each sentence subjected to perturbations. Then, we evaluate the translation quality for each noise level using overlap and neural reference based metrics.

\section{{\mtlens} UI}

The web user interface is organized into four main sections, each corresponding to a different MT task (Figure \ref{fig:app_overview}). It is built in Python using the Streamlit framework\footnote{\href{https://streamlit.io/}{https://streamlit.io/}}. In this section, we describe the tools implemented in the user interface of {\mtlens} and demonstrate its utility by evaluating two state-of-the-art NMT systems: madlad-400-3B \cite{kudugunta2024madlad} and NLLB-3.3B \cite{nllbteam2022language}, for the Catalan-to-English translation direction.

\begin{figure}[t]
\centering
  \includegraphics[width=1\linewidth]{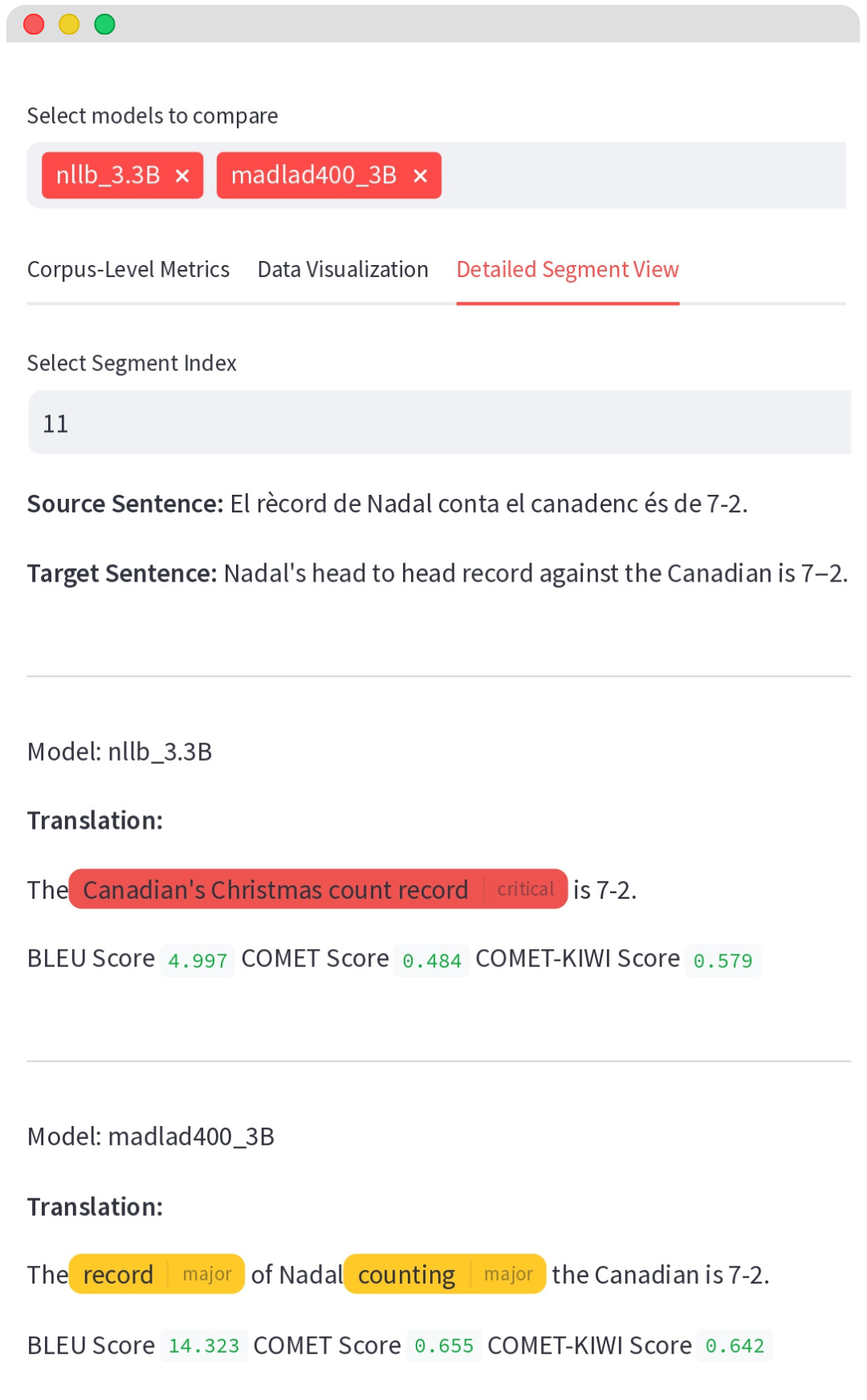}
  \caption {Segment comparison with error spans produced by madlad-400-3B and NLLB-3.3B systems.}\label{fig:seg2seg}
\end{figure}

\subsection{{\normalfont{\mtlens} UI:} Translation}

\subsubsection{Segment-by-Segment Comparison}

\mtlens\ allows users to analyze and compare translations across different systems. It first displays the source and target sentences for the selected segment, followed by the corresponding translations from the selected models (Figure \ref{fig:seg2seg}).

Identifying and categorizing the errors made by NMT systems can be highly informative when comparing different models. In {\mtlens}, if the {\xcomet} metric has been computed when evaluating a model, we use it to highlight error spans in a translation and marking them with different colors that indicate the severity of the error: red for critical errors, yellow for major errors, and blue for minor errors. We also provide individual segment scores for {\bleu}, {\comet}, and {\cometkiwi}, which can be used to understand the translation's similarity to the reference text, its semantic similarity to the reference, and its semantic similarity to the source sentence respectively.

In Figure \ref{fig:seg2seg}, we show an example of the segment-by-segment comparison page. We can see that the translation given by NLLB-3.3B has been categorized as critical by {\xcomet} while madlad-400-3B attains better results in individual metric scores, although it still produces two major errors.

\begin{figure*}[t]
\centering
  \includegraphics[width=0.95\linewidth]{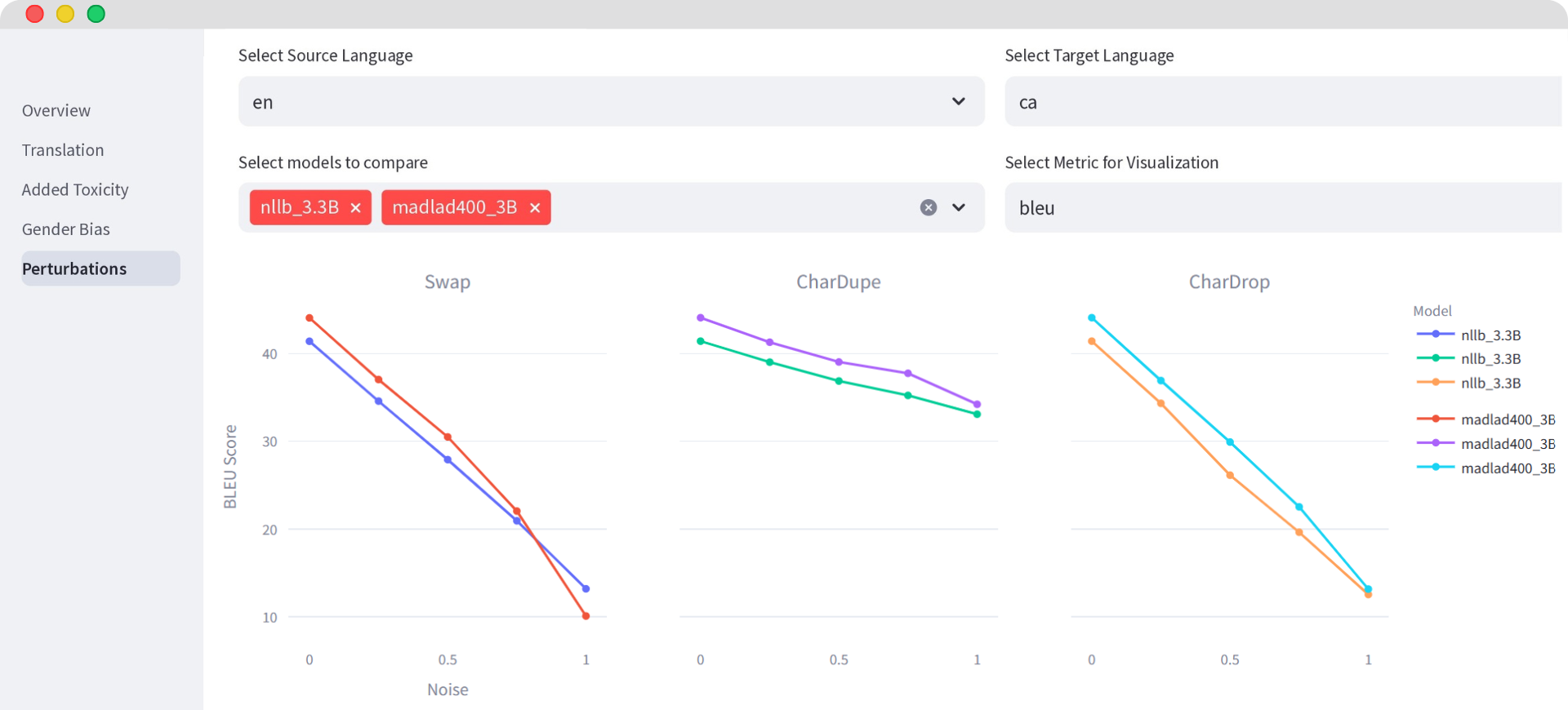}
  \caption {An image from the Perturbations page in the {\mtlens} UI. Users can navigate between the following options: (1) Overview, (2) Translation, (3) Added Toxicity, (4) Gender Bias, and (5) Perturbations. }\label{fig:app_overview}
\end{figure*}

\subsubsection{Segment-Length Analysis}

Sentence length significantly influences system performance, with NMT systems often producing lower-quality translations for very long sentences \cite{koehn-knowles-2017-six}. For analyzing the effect of sentence length on translation quality, {\mtlens} offers interactive scatter plots, where the x-axis represents the number of words in a sentence, and the y-axis displays the corresponding score of that sentence for the selected metric.

\subsubsection{Statistical Significance Testing}

When comparing NMT systems, {\mtlens} provides a visual interface for computing statistical significance testing through bootstrapped t-tests \cite{koehn-2004-statistical} on {\bleu}, {\comet} and {\cometkiwi} metrics. Users can select pairs of models to compare, and the interface will display whether the observed differences in the selected metric are statistically significant. 

\subsection{{\normalfont{\mtlens}  UI:} Added Toxicity}

Inspecting those segments that contain added toxicity can be particularly interesting when evaluating a NMT system using \textsc{HolisticBias}. Using {\mtlens}  UI, the user can see the obtained metrics aggregated at the system level and inspect the terms detected by {\etox} for the selected model along a specific axis at the segment level.

\subsection{{\normalfont{\mtlens}  UI:} Gender Bias}

Evaluating gender bias is crucial for developing fair and inclusive NMT systems. The {\mtlens} UI offers a dedicated interface for assessing gender bias, showing aggregated metrics at the system level for the selected dataset. This interface is organized into two tabs, each corresponding to \textsc{MuST-SHE} and \textsc{MMHB}, respectively.

\subsection{{\normalfont{\mtlens} UI:} Perturbations}

Understanding how different translation systems handle input perturbations is crucial for assessing their robustness to real-world applications. {\mtlens} UI, provides different visualizations to compare system performance for each type of noise evaluated. In Figure \ref{fig:app_overview}, we present an example of the Perturbations interface. The results show that madlad-400-3B exhibits greater robustness than NLLB-3.3B across all types of synthetic noise evaluated using the {\bleu} metric.

\section{Conclusion}

In this paper, we introduced {\mtlens}, a framework designed to address existing gaps in MT evaluation by unifying various evaluation strategies. {\mtlens} supports a diverse range of MT tasks, including traditional translation quality evaluation, gender bias detection, added toxicity, and robustness to character noise. By building upon the widely adopted LM-eval-harness library, {\mtlens} provides seamless integration for evaluating both NMT and LLM-based models across various tasks. {\mtlens} also offers a platform designed to provide insights into system performance. We believe {\mtlens} has the potential to become the new adopted framework for evaluating NMT systems in the research community.

\section{Limitations}

Our tool is designed to provide a robust framework for researchers in machine translation to analyze various aspects of evaluation with ease. It is built to be adaptable and extendable, enabling the community to seamlessly incorporate new machine translation datasets. While standard evaluation metrics for machine translation are consistent across datasets, metrics for analyzing specific phenomena like gender bias and toxicity often require customization to suit the dataset. As a result, incorporating new metrics or datasets might occasionally require some additional effort or minor adjustments to the user interface. However, this is an area of active development, and we aim to implement methods in the future that will enhance flexibility and streamline these tasks even further.

\section{Ethical Statement}
Gender bias and toxicity in machine translation are multifaceted challenges that encompass a wide range of phenomena. The gender bias datasets used in this work primarily focus on evaluating coreference accuracy within a binary classification framework (male/female). For toxicity detection, we rely on the ETOX dataset, which identifies content deemed universally toxic, independent of context. While we acknowledge the limitations of these approaches, our objective is to represent widely recognized datasets for these tasks and contribute to a broader understanding of machine translation evaluation. This work does not aim to provide an exhaustive treatment of these complex issues but rather to offer a representative perspective.

\section{Acknowledgements}

This work has been promoted and financed by the
Generalitat de Catalunya through the Aina Project.\\

\noindent
This work has been supported by the Spanish
project PID2021-123988OB-C33 funded by MCIN/AEI/10.13039/501100011033/FEDER, UE.\\

\noindent
This work is partially supported by DeepR3 (TED2021-130295B-C32) funded by MCIN/AEI/10.13039/501100011033 and European Union NextGeneration EU/PRTR.\\

\noindent
This work was supported by EU Horizon 2020 project ELOQUENCE13\footnote{\url{https://eloquenceai.eu/}} (grant number 101070558).

\bibliography{latex/acl_latex}

\appendix

\end{document}

%% file: latex/tables/datasets.tex
\begin{table*}[ht]
\centering
\small
\begin{tabularx}{\textwidth}{lllr}
\toprule
\textbf{Task} & \textbf{Dataset} & \textbf{Task name} &  \textbf{Languages}  \\
\midrule
General-MT & \floresdataset \cite{nllbteam2022language} & \{src\}\_\{tgt\}\_flores\_\{split\} & 200\\

& \textsc{NTREX-128} \cite{federmann-etal-2022-ntrex} & \{src\}\_\{tgt\}\_ntrex & 128\\

& \textsc{TATOEBA} \cite{tiedemann-2020-tatoeba} & \{src\}\_\{tgt\}\_tatoeba & 555 \\
& \textsc{NTEU} \cite{bie-etal-2020-neural} & \{src\}\_\{tgt\}\_nteu & 25 \\
\cdashlinelr{1-4}
Added Toxicity & \holisticbias \cite{smith-etal-2022-im} & \{src\}\_\{tgt\}\_\{axis\}\_hb & 1  \\
\cdashlinelr{1-4}
Gender Bias & \textsc{MuST-SHE} \cite{bentivogli-etal-2020-gender} & \{src\}\_\{tgt\}\_must\_she & 5  \\
& \textsc{MMHB} \cite{tan2024massivemultilingualholisticbias} & \{src\}\_\{tgt\}\_mmhb\_\{split\} & 7  \\
& \textsc{MT-GenEval} \cite{currey-etal-2022-mtgeneval} & \{src\}\_\{tgt\}\_geneval\_\{split\} &  8 \\
\cdashlinelr{1-4}
{\footnotesize Robustness to Character Noise} & \floresdataset devtest & \{src\}\_\{tgt\}\_perturbations & 200  \\
\bottomrule
\end{tabularx}
\caption{Machine translation datasets natively supported by {\mtlens} grouped by task type.}\label{table:datasets}
\end{table*}

%% file: latex/tables/metrics.tex
\begin{table*}[ht]
\centering
\small
\begin{tabularx}{\textwidth}{XXX}
\toprule
\textbf{Type} & \textbf{Name} & \textbf{Implementation} \\
\midrule
Overlap Reference-based & \bleu\; \cite{papineni-etal-2002-bleu} & \texttt{SacreBLEU} \cite{post-2018-call} \\
& \ter\; \cite{snover-etal-2006-study} & \texttt{SacreBLEU} \\
& \chrf\; \cite{popovic-2015-chrf} & \texttt{SacreBLEU} \\
\cdashlinelr{1-3}
Neural Reference-based & \comet\; \cite{rei-etal-2022-comet} & \texttt{unbabel-comet} \cite{stewart-etal-2020-comet} \\
& \bleurt\; \cite{sellam-etal-2020-bleurt} & \texttt{transformers} \cite{wolf-etal-2020-transformers} \\
& \metricx & \texttt{metricx} \cite{juraska-etal-2023-metricx} \\
\cdashlinelr{1-3}
Neural Reference-based and error span & \xcomet\; \cite{guerreiro2023xcomet} & \texttt{unbabel-comet} \\
\cdashlinelr{1-3}
Quality estimation & \metricxqe\; & \texttt{metricx} \\
& \cometkiwi\; \cite{rei-etal-2022-cometkiwi} & \texttt{unbabel-comet} \\
\cdashlinelr{1-3}
Quality estimation and error span & \xcometqe\; & \texttt{unbabel-comet} \\
\cdashlinelr{1-3}
Word lists & {\etox} & \texttt{nllb} \cite{nllbteam2022language} \\
\cdashlinelr{1-3}
 Embedding-based & {\mutox} \cite{costa-jussa-etal-2024-mutox} & \texttt{seamless} \cite{communication2023seamlessmultilingualexpressivestreaming} \\
& {\detoxify} & \texttt{detoxify} \cite{detoxify} \\
\bottomrule
\end{tabularx}
\caption{Evaluation metrics supported by \mtlens.}\label{table:metrics}
\end{table*}